\documentclass{INTERSPEECH2023}

\interspeechcameraready 
\usepackage[final]{pdfpages}
\usepackage{graphicx}
\usepackage{blindtext}
\usepackage{hyperref} 
\usepackage{svg}
\usepackage{multirow}
\usepackage{arabtex}
\usepackage{utf8}
\setcode{utf8}

\newcommand\blfootnote[1]{%
  \begingroup
  \renewcommand\thefootnote{}\footnote{#1}%
  \addtocounter{footnote}{-1}%
  \endgroup
}

\title{\textit{N}-Shot Benchmarking of Whisper on Diverse Arabic Speech Recognition}
\name{Bashar Talafha$^{1*}$, Abdul Waheed
$^{2 *}$, Muhammad Abdul-Mageed
$^{1, 2}$}

\address{
  $^1$The University of British Columbia, Canada\\
  $^2$Mohamed bin Zayed University of Artificial Intelligence, UAE }
\email{\{btalafha@mail.,muhammad.mageed@\}ubc.ca,abdul.waheed@mbzuai.ac.ae}

\begin{document}

\maketitle
 
\begin{abstract}
Whisper, the recently developed multilingual weakly supervised model, is reported to perform well on multiple speech recognition benchmarks in both monolingual and multilingual settings. However, it is not clear how Whisper would fare under \textit{diverse} conditions even on languages it was evaluated on such as Arabic. In this work, we address this gap by comprehensively evaluating Whisper on several varieties of Arabic speech for the ASR task. Our evaluation covers most publicly available Arabic speech data and is performed under \textit{n}-shot (zero-, few-, and full) finetuning. We also investigate the robustness of Whisper under completely novel conditions, such as in dialect-accented standard Arabic and in unseen dialects for which we develop evaluation data. Our experiments show that although Whisper zero-shot outperforms fully finetuned XLS-R models on all datasets, its performance deteriorates significantly in the zero-shot setting for five unseen dialects (i.e., Algeria, Jordan, Palestine, UAE, and Yemen).  
\end{abstract}
\noindent\textbf{Index Terms}: Arabic, automatic speech recognition, Arabic dialects, Whisper, speech analysis, natural language processing, speech technology.

\blfootnote{$^*$Equal contribution}

\section{Introduction} \label{intro}
    Self-supervised and weakly-supervised training paradigms that exploit massive amounts of data have recently resulted in impressive performance improvements on a wide range of tasks across modalities~\cite{liu2022audio, schiappa2022self, qiu2020pre, kim22kinterspeech}. One such example is \textit{Whisper} \cite{radford2022robust}, which is a multilingual multi-task weakly supervised speech model. Although Whisper was evaluated on multiple speech benchmarks, often demonstrating good performance, it remains unclear how it would fare in scenarios with significant \textit{speech variability}. In this work, we investigate the generalization capability of Whisper on various Arabic dialects and accents for the speech recognition task. Arabic is an excellent context for testing the robustness of Whisper, due to its diverse collection of languages and dialectal varieties. In particular, rather than being a single monolithic language, Arabic is a \textit{collection} of languages and dialectal varieties  that vary extensively over \textit{geography} (the Arab world extends across Asia and Africa). Arabic is also \textit{diaglossic}, with a so-called \textit{high} variety (Modern Standard Arabic [MSA]) that is spoken sometimes in education and government and several \textit{low} varieties (dialects) that are used in everyday communication. Due to this rich sociolinguistic context and variability, high performance on MSA speech can not realistically guarantee comparable performance on dialects or even accented MSA.
    
In this paper, we test Whisper within this context of wide variability. In particular, we benchmark Whisper on an extensive number of existing and new Arabic datasets under \textit{n}-shot conditions (i.e., zero- and few-shot as well as full finetuning). We use XLS-R \cite{babu2021xls} as our baseline and evaluate the generalization capability of Whisper on both \textit{out-of-domain} datasets as well as \textit{unseen dialects} that have not been studied before. Moreover, we \textit{stress-test} Whisper models  by examining robustness of their finetuned versions on unseen dialects and dialect-accented MSA.

\section{Related Work} \label{related-work}
End-to-end (E2E) deep learning models have shown significant improvements in ASR performance by learning directly from the audio waveform without relying on an intermediate feature extraction layer \cite{wang2019overview}. One recent example of an E2E ASR is the recently proposed Whisper~\cite{radford2022robust}, a transformer-based sequence-to-sequence model trained in a multitask fashion. Namely, Whisper is trained on ASR, voice activity detection, language identification, and speech translation. It is trained in a weakly supervised manner with up to $680$K hours of labeled audio data. The model is tested in zero-shot settings on multiple datasets and achieves state-of-the-art performance on benchmark datasets such as librispeech~\cite{panayotov2015librispeech}, TEDLIUM~\cite{rousseau2012ted}, and Common Voice~\cite{ardila2019common}. Although Whisper displays robustness on these benchmark datasets, recent research has demonstrated that it can be vulnerable to adversarial noise. This vulnerability can lead to significant degradation in performance and hurt model ability to transcribe targeted sentences \cite{olivier2022there}. Therefore, it remains unclear how Whisper would fare in scenarios with significant variability such as on Arabic. Although the pretraining data of Whisper has $\sim739$ hours of Arabic speech, it is unclear what varieties of Arabic this data covers.

The E2E model proposed by~\cite{ahmed2019end} is the first to introduce a lexicon-free approach for Arabic ASR using recurrent neural networks with connectionist temporal classification (CTC). DeepSpeech E2E is trained by~\cite{belinkov2019analyzing} on Arabic and English datasets. These authors also  investigate the internal representations learned by the model on different speech tasks. An E2E transformer-based architecture is also proposed by~\cite{hussein2022arabic}, employing a multitask CTC/attention objective function. The model achieves state-of-the-art on both Modern Standard Arabic (MSA) and dialects. 

While E2E ASR models can streamline the ASR process and outperform traditional models, they still demand a significant amount of labeled data for training from scratch. This poses a challenge  for low-resource languages such as Arabic. To address this issue, self-supervised and semi-supervised models have gained popularity. The reason is that these models learn useful representations from large amounts of unlabeled or weakly-labeled data and can be finetuned on different speech tasks~\cite{baevski2020Wav2vec,babu2021xls,hsu2021hubert,chung2021w2v}. Similar to text- and image-based pretrained models, these models are used in two stages: pretraining, where the model learns the representation, and finetuning, where these learned representations are used for a specific task. 
Wav2vec2.0 \cite{baevski2020Wav2vec} is a pretraining model that is self-supervised and is based on a combination of convolutional neural networks (CNNs) and transformers that learn to predict a set of masked audio input samples. An extension of Wav2vec2.0 is XLS-R \cite{babu22interspeech}, which is a cross-lingual pretraining model trained on $436$k hours of speech from $128$ languages (including $95$ Arabic speech hours from common voice \cite{ardila2019common} and VoxLingua107 \cite{valk2021voxlingua107}). One work finetunes XLS-R on Arabic data from common voice 6.1~\cite{bakheet2021improving}.  

Another self-supervised framework is w2v-BERT~\cite{chung2021w2v}, which combines contrastive learning and masked language modeling (MLM).
w2v-BERT was adapted for Arabic ASR by finetuning on FLEURS dataset~\cite{conneau2022FLEURS}. The Arabic subset of FLEURS represents dialect-accented standard Arabic spoken by Egyptian speakers, though it has not been extensively studied or presented as a noteworthy example of dialect-accented Arabic.
Unlike Whisper, an issue of Wav2vec2.0 and w2v-BERT is that these require a finetuning stage as they lack proper decoding. 

\section{Datasets and Preprocessing}\label{sec:datasets_preproc} \label{dataset_preprocessing}
We make use of a wide range of Arabic datasets including data covering MSA, various Arabic dialects, and accented MSA. Each of these datasets provides a unique perspective on the challenges and complexities of Arabic ASR. We introduce these datasets next.
\\
\newline
\noindent
\textbf{Common Voice \cite{ardila2019common} (v6.1, v9.0, v11.0).} These datasets are from Mozilla Common Voice,\footnote{https://commonvoice.mozilla.org/en/about.} where volunteers record sentences in \textit{MSA} with each recording validated by at least two users. We exploit three versions of common voice: v6.1, v9.0, and v11.0. These have $50$, $88$, and $89$ hours, respectively. 
\\
\newline
\noindent
\textbf{MGB-2 \cite{ali2016mgb}.} This dataset contains about $1,200$ hours of Arabic broadcast data from Aljazeera Arabic TV channel collected over $10$ years. It includes time-aligned transcription obtained from light-supervised alignment. The dataset has $70$\% \textit{MSA} while the remaining $30$\% is in various Arabic \textit{dialects}.
\\
\newline
\noindent
\textbf{MGB-3~\cite{ali2017speech}.} This dataset contains $80$ programs from YouTube channels based in \textit{Egypt}, featuring various genres like comedy, cooking, sports, and science talks. The dataset includes transcriptions of the first $12$ minutes of each program, resulting in a total of $4.8$ hours of transcribed data.
\\
\newline
\noindent
\textbf{MGB-5 \cite{ali2019mgb}.} This dataset contains $10.2$ hours of \textit{Moroccan} Arabic speech data from $93$ YouTube videos belonging to seven genres such as comedy, cooking, and sports.
\\
\newline
\noindent
\textbf{FLEURS \cite{conneau2022FLEURS}.} This contains a subset of Arabic language that represents \textit{standard Arabic spoken with an Egyptian accent}. We utilize this subset to evaluate the robustness of our MSA models in accented conditions (section \ref{robustness}). The dataset contains $4.39$ hours of \textit{MSA produced by Egyptian native speakers}.
\\
\newline
\noindent
\textbf{AraYouTube.} We introduce this new dataset for our work. We manually identify soap opera videos in \textit{Algerian}, \textit{Jordanian}, \textit{Palestinian}, \textit{UAE}, and \textit{Yemeni} dialects and task a team of trained native speaker annotators to transcribe them. We acquire $3.2$, $0.98$, $4.4$, and $2.94$ hours of the dialects, respectively. We use AraYouTube as unseen datasets solely for evaluation purposes (section \ref{in-house}).
\\
\newline
\noindent
\textbf{Preprocessing.} Some of the datasets we use have inconsistencies. For example, in CV6.1 the utterance \RL{فَقَالَ لَهُمْ} (faqaAla lahumo) has complete diacritic markings, while the utterance \RL{فإذا النجوم طمست} (f$<$*A Alnjwm Tmst) does not have any diacritics, even though both come from the Quran. For this reason, we 
 follow~\cite{chowdhury21interspeech} in standardizing the data. Namely, we (a) remove all punctuation except the \% and @ symbols; (b) remove diacritics, Hamzas, and Maddas; and (c) transliterate all eastern Arabic numerals to western Arabic numerals (e.g. \RL{٩٢} to 29). Since we do not treat code-switching in this work, we remove all Latin characters.

\section{Experiments} \label{experiments}

\begin{table*}[t!]
    \caption{ \label{tab:main-result}
    Results on Test in Word Error Rate (WER \big\downarrow~) and Character Error Rate (CER \big\downarrow~) (/ separated) results for zero-shot, few-shot (in hours), and full finetuned (last four rows) models. \textbf{-}: not applicable; \textbf{NA}: model did not converge.}
    \label{tab:all-results}
    \begin{tabular}{cllllllll}
    \toprule
    \multicolumn{1}{c}{\textbf{Setting}}  & \multicolumn{1}{c}{\textbf{CV6.1}} & \multicolumn{1}{c}{\textbf{CV9.0}} & \multicolumn{1}{c}{\textbf{CV11.0}} & \multicolumn{1}{c}{\textbf{MGB-2}} & \multicolumn{1}{c}{\textbf{MGB-3}} & \multicolumn{1}{c}{\textbf{MGB-5}} & \multicolumn{1}{c}{\textbf{FLEURS}} \\
    \midrule
    \multirow{1}{*}{Zero-shot} & 15.93 / 5.69 & 19.2 / 6.59 & 19.41 / 6.76 & 34.74 / 17.75 & 43.53 / 21.87 & 82.97 / 49.12 & 11.56 / 3.74 \\
    
    \midrule
    \multirow{1}{*}{1h} & 15.65 / 4.52 & 18.45 / 5.81 & 17.98 / 5.3 & 21.93 / 10.29 & 34.23 / 14.14 & 61.46 / 26.71 & 12.64 / 3.95 \\
    
    \multirow{1}{*}{2h} &  14.27 / 4.3 & 16.92 / 5.05 & 16.55 / 4.92 & 25.02 / 12.57 & 33.26 / 14.26 & 59.58 / 25.76 & 11.32 / 3.64 \\

    \multirow{1}{*}{4h} &  12.85 / 3.77 & 16.15 / 4.92 & 15.47 / 4.72 & 24.43 / 12.54 & 32.02 / 13.39 & 56.24 / 23.41 & 11.09 / 3.98 \\

    \multirow{1}{*}{8h} &  12.03 / 3.5 & 14.89 / 4.43 & 14.9 / 4.59 & 21.69 / 11.11  & \multicolumn{1}{c}{-} & 54.98 / \textbf{22.93} &  \multicolumn{1}{c}{-} \\

    \multirow{1}{*}{16h} & 10.95 / 3.23 & 14.44 / 4.33 & 13.92 / 4.31 & 21.57 / 11.22 & \multicolumn{1}{c}{-} & \multicolumn{1}{c}{-} & \multicolumn{1}{c}{-} \\

    \midrule

    \multirow{1}{*}{XLS-R}
    & 30.32 / 9.33 & 32.35 / 9.89 & 31.16 / 9.35 & \multicolumn{1}{c}{NA} & 55.12 / 21.06 & \multicolumn{1}{c}{NA} & \multicolumn{1}{c}{NA} \\

    \multirow{1}{*}{XLS-R-LM}
     & 21.62 / 7.19 & 23.4 / 7.988 & 22.56 / 7.55 & \multicolumn{1}{c}{NA} & 48.41 / 19.6 & \multicolumn{1}{c}{NA} & \multicolumn{1}{c}{NA} \\

    \multirow{1}{*}{Whisper\textsubscript{Small}}  & 22.21 / 7.09 & 24.61 / 7.94 & 24.12 / 7.93 & 29.66 / 14.44 & 49.68 / 22.48 & 73.98 / 33.46  & 23.76 / 9 \\

    \multirow{1}{*}{Whisper\textsubscript{Large}}  & \textbf{10.81} / \textbf{3.24} & \textbf{12.97} / \textbf{4.18} & \textbf{13.28} / \textbf{4.23} & \textbf{15.49} / \textbf{8.62} & \textbf{31.39} / \textbf{13.25} & \textbf{53.82} / 22.99 & \textbf{10.36} / \textbf{3.3} \\

    \toprule 
    \end{tabular}
\end{table*}

We experiment with two versions of Whisper (\textit{large-v2} and \textit{small}) on all the datasets that we consider in our study. As stated earlier, our main objective is to investigate Whisper's robustness under dialectal and accented conditions. We evaluate Whisper in three settings: zero-shot, few-shot, and full finetuning. For comparison, we also finetune the XLS-R model.

\subsection{Experimental Setup}
We sample audio with $16$kHz rate and perform preprocessing steps on the text described in Section~\ref{sec:datasets_preproc}. We use transformers\footnote{https://huggingface.co/docs/transformers/index} for the training and evaluation pipeline. We use single node $4$xV100 - 32GB and $4$xA100-40GB GPUs for all of our experiments. The \textit{whisper-large-v2}\footnote{https://github.com/openai/whisperavailable-models-and-languages} does not fit into a single V100 GPU even with batch size 1. To overcome this, we parallelize the model across the GPUs enabled by deepspeed\footnote{https://www.deepspeed.ai/tutorials/zero/} ZeRo stage-2. To fully utilize the GPU memory, we use single device batch size $32$ and effective batch size $256$ including gradient accumulation steps (varying subject to dataset size) for multi-GPU training. During finetuning (applied to both few-shot and full finetuning), the feature extraction layer is frozen (for both XLS-R and Whisper models) while the other layers are trainable. This is because the feature extraction layer has already learned the general representations of speech signals during the pretraining process. For optimization, we use AdamW~\cite{loshchilov2017decoupled} with a learning rate 1e-5 and 3e-4 (for Whisper and XLS-R respectively) and warmup steps at $500$. The remaining optimizer's parameters are the default. We train each model and dataset split configuration for $100$ epochs with early stopping patience at $3$ and threshold $1$. We find that training converges way before the full $100$ epochs. For decoding, we use max length $225$ and we do not apply any processing steps on decoded outputs. 

\subsection{{\it N}-shot Learning} \label{zsfsl}

We perform all of our experiments with Whisper models (large-v2 and small) as well as with XLS-R on all the publicly available Arabic ASR datasets described in Section~\ref{sec:datasets_preproc}. We do not find any fully supervised or zero/few-shot model for Arabic ASR evaluated on all these datasets to which we can compare. As pointed out in Section~\ref{sec:datasets_preproc}, these datasets vary in terms of dialect, accent, and conditions\footnote{we call them conditions because across the datasets not only dialect or accent changes but other factors such as background noise, native sampling rate, etc also vary}. We perform zero-shot, and few-shot evaluations on whisper-large-v2, and we fully fine-tune XLS-R and whisper models (large-v2, small). 
For zero-shot evaluation, we employ the processing steps stated in Section~\ref{sec:datasets_preproc} on decoded output and ground truth before computing the Word Error Rate (WER) and Character Error Rate (CER).

\noindent 
\textbf{Zero-shot evaluation.} Whisper performs quite well in the zero-shot setting on a wide range of speech tasks including, but not limited to, the ASR task. We evaluate Whisper (large-v2, 1.5B) in a zero-shot setting on all the datasets described in 
Section~\ref{sec:datasets_preproc}. We report WER and CER on test sets. We apply preprocessing techniques in two ways:
(1) we apply all the preprocessing steps mentioned in Section~\ref{sec:datasets_preproc} except removing diacritics, (2) we apply all the preprocessing steps including removing the diacritics. We observe that without removing the diacritics, zero-shot results improve by almost $10$ points in terms of WER/CER across different datasets. We also observe that the difference is quite big when we remove the diacritics, particularly for common voice datasets. Upon inspection, we observe that common voice datasets have speech based on historical textbooks which are heavily diacritised while FLEURS and others consist of MSA comparatively less diacritized. Zero-shot results are stated in Table~\ref{tab:all-results}.\\
\noindent 
\textbf{Few-shot finetuning.}
For few-shot finetuning, we take \textit{whisper-large-v2}\footnote{https://huggingface.co/openai/whisper-large-v2} checkpoint using the same objective the original Whisper is trained with. We split the data as per $2^n, n=0, 1..,where 2^n <=min(16, train)$. For each split, we train Whisper independently (meaning we do not use the checkpoint from the previous split). We apply our preprocessing steps and use an Arabic tokenizer after processing the text. Namely, we apply the respective tokenizer used by each tool. In the case of XLS-R, words are split into characters while BPE is employed in the case of Whisper. We report WER and CER on the Dev and Test splits of each dataset. The \textit{Wav2vec2.0 XLS-R} \footnote{https://huggingface.co/facebook/Wav2vec2-XLS-R-300m} model is pretrained to learn speech representations, in contrast to the Whisper model which is pretrained to learn transcriptions that, unlike XLS-R, empowers Whisper to be used without any finetuning. Additionally, we do not anticipate the XLS-R model to perform well on few-shot learning, which involves training on a small amount of labeled data. Therefore, we report only full finetuning for the XLS-R model. We conduct additional tests by integrating a Kenlm-based n-gram language model \footnote{https://github.com/kpu/kenlm} with $n=3$. In the case of MSA datasets, language model is generated using a combination of the training subsets of each MSA dataset. For MGB-3 and MGB-5, only the training set of each dataset was utilized to create a dialect-based language model. We refer to this ASR model as XLS-R-LM.

\noindent 
\textbf{Full finetuning (FFT).} To perform full finetuning, we use the same full training split of each dataset and apply the same preprocessing techniques as in the zero-shot and few-shot evaluations.
We notice that Whisper models, when compared to the XLS-R models, exhibit superior performance across all datasets. However, we observe that the XLS-R model with default hyperparameters fails to converge on MGB-5, likely due to the considerable dissimilarity between the Moroccan dialect and the MSA used to pretrain XLS-R. Moreover, we observe that the default hyperparameters are not optimal for FLEURS and MGB-2. This indicates that more exploration of hyperparameters is necessary. In comparison, Whisper models remain robust and converge effectively on these datasets (with particularly outstanding results achieved on the FLEURS dataset).
Furthermore, in terms of full finetuning, we notice that for Whisper the results obtained across all datasets are consistently superior to those obtained through zero-shot and few-shot evaluations (in terms of both WER and CER). This observation highlights the potential of full finetuning as an effective strategy for enhancing overall performance, particularly when applied to the Whisper architecture.

\section{Results}\label{results}
\noindent
\textbf{Zero-shot results.} In zero-shot evaluation, we observe that scaling, in terms of architecture, yields better results and often is on par with the fully finetuned small model. Although it is not a fair comparison since large-v2 is about x5 bigger than the whisper-small. But to put things in perspective, the WER for large-v2 on the FLEURS test set is $11.56$ compared to the fully finetuned small WER of $23.76$ on the same test. From our experiments, we notice that while whisper does quite well on standard benchmark datasets and surprisingly even on accented conditions such as FLEURS, it fails to generalize on dialectal Arabic speech in a zero-shot setting. However, performance in accented conditions needs to be scrutinized further as FLEURS has very small evaluation sets.

\noindent\textbf{Few-shot and full finetuning results.}
In few-shot finetuning, we observe that up to $4$ hours of the training data yield on-par performance compared to full finetuning in most cases. For example, training whisper-large-v2 on $4$ hours of the CV11.0 training set gives $15.47$ WER compared to the full finetuned model (which is $13.28$) when full training has almost $32$ hours of speech data. When we do few-shot finetuning of Whisper models on MGB-2 training set, we find that adding more training hours sampled from whole training may not aid in getting better results neither on Test nor Dev sets. We hypothesize that this is mainly due to the fact that MGB-2 training set has multiple dialectal variations. For few-shot finetuning, the sampled data distribution (ie: dialect) may or may not be the same as test and validation sets. This also corroborates our finding that Whisper's performance degrades in unseen and novel conditions. Further, we finetune the XLS-R model on all the datasets and when we incorporate a statistical language model during decoding in XLS-R it consistently improves performance across all datasets, resulting in a decrease of nearly 9 WER. 
From our experiments, we observe that full finetuning results for the standard benchmark are close to the human baseline in terms of WER but still far for dialectal and accented speech. For comparison, training Whisper on $10.2$ hours of MGB-3 obtains WER of $53.82$ while $16$ hours from CV11.0 results in $13.92$ WER on respective test sets.

\vspace{-1mm}
\section{Robustness on MSA-Accented Conditions} \label{robustness}
\vspace{-1mm}
Whisper is trained on roughly $680,000$ hours of speech data. Unlike general representation models such as XLS-R, Whisper is trained to perform downstream speech tasks without any supervised training. The Whisper pretraining data has roughly $17$\% of non-English speech including over $700$ hours and $2,300$ hours of Arabic speech data for recognition and translation tasks, respectively. It is not completely clear what all Arabic speech datasets were included (except that we know CV9.0 and FLEURS are part of these data). As expected, in our evaluation, Whisper performs quite well on CV9.0 and FLEURS test and validation sets. Robustness of the Arabic ASR models trained on MSA, however, have not been examined under various conditions such as dialectal and accented Arabic speech. To fill this gap, we evaluate and further train MSA models on dialectal and accented conditions. We first finetune Whisper-large-v2 on CV11.0 (MSA) and evaluate it on various dialects and accented speech. As seen in Table~\ref{tab:robustness}, we observe that the fully fine-tuned MSA model (referring to Whisper-large-v2 finetuned on CV11.0) does worse on unseen dialects and accented speech than the zero-shot in almost all scenarios. For example, the WER of the MSA model (Whisper-large-v2 finetuned on CV11.0) on the MGB-3 test set is $55.31$ compared to the zero-shot WER of $31.39$ on the same test set. 
To investigate the generalization and adaptive capability of Arabic MSA models further on unseen dialects, accents, and conditions, we adapt models (by continued finetuning) as recommended by~\cite{ali2019mgb, ali2017speech}. We evaluate these adapted models on the same distribution (condition) as well as again on MSA data as a zero-shot model. We observe that the adapted model (i.e., Whisper-zero-shot $\rightarrow$ CV11.0 $\rightarrow$ FLEURS) is just on par or often even worse compared to the zero-shot and fully fine-tuned model (i.e., Whisper-zero-shot $\rightarrow$ FLEURS). More specifically, the WER on the CV11.0 test set of the MSA model finetuned on MGB-3 (Egyptian) is $22.43$ compared to the MSA model WER of $13.28$ and zero-shot WER $15.93$ on the same test set. 

\section{Performance on Novel Dialects} \label{in-house}
To investigate robustness of Whisper on novel conditions, we further perform zero-shot evaluation on novel dialects collected and annotated by a group of native speakers as described in Section~\ref{sec:datasets_preproc}. We report WER and CER on five novel dialects, which we hypothesize may not have been part of Whisper training data. From our evaluation, we find that Whisper-large-v2 on standard benchmarks such as FLEURS is close to the human baseline ($4$\% WER) in the zero-shot settings but, as shown in Table 3, is not able to generalize well on completely novel and unseen conditions. Upon inspecting the decoded output, we find that the model generates random and repetitive sentences. For example, we find the phrase \< اشتركوا في القناة> $206$ times in UAE and $152$ times in Palestinian decoded output. We believe this is a result of pretraining data leakage. While, for Palestine and Jordan dialects, Whisper does reasonably  get the best WER although not without noise in its output. We hypothesize that this lower WER is due to Jordan and Palestine sharing more vocabulary with MSA than the other dialects. We conclude that without finetuning, Whisper fails on all unseen dialects.

\begin{table}[h]
\caption{Robustness Test. The base model of these experiments is trained on CV11. Results under Dev and Test represent WER/CER.}
\begin{tabular}{cccc}
    \toprule
\textbf{Adapt}  & \textbf{Dataset} &\textbf{ Dev} & \textbf{Test }\\
    \midrule

None   & CV11    & 8.69 / 2.45   & 13.28 / 4.23   \\
                       None   & FLEURS  & 16.97 / 5.43  & 16.85 / 5.69   \\
                       FLEURS & FLEURS  & 10.24 / 3.02  & 10.16 / 3.44   \\
                       FLEURS & CV11    & 11.35 / 3.53  & 15.15 / 4.90 \\ 
                       None & MGB-3    & 53.23	/ 26.97  & 55.31	/ 28.49 \\
                       MGB-3 & MGB-3    & 31.47 / 12.52  & 31.59	/ 13.33 \\
                       MGB-3 & CV11    & 18.16	/ 5.34  & 22.43 / 6.91 \\
    \bottomrule
\end{tabular} \label{tab:robustness}
\end{table}

\begin{table}[h]
  \caption{Statistics and evaluation results of the AraYouTube dataset for different dialects (as an unseen condition). }
  \label{tab:word_styles}
  \centering
  \begin{tabular}{lccc}
    \toprule
    \textbf{Dialect} & \textbf{Hours} & \textbf{Segments} & \textbf{WER/CER} \\
    \midrule
    Algeria & 0.9 & 840  & 103.44 / 81.94 \\
    Jordan & 1.20 & 1000  & 72.80 / 58.95 \\
    Palestine & 1.6 & 1,111  & 51.92 / 19.42 \\
    UAE & 2.37 & 2,000 & 102.83 / 83.48 \\
    Yemen & 1.27 & 1000 & 102.66 / 81.26 \\
    \bottomrule
  \end{tabular}
\end{table}

\section{Conclusion} \label{conclusion}
We benchmark Whisper models on Arabic ASR for a wide range of dialects and conditions. Our empirical investigations allow us to observe that while Whisper is a robust and strong \textit{n}-shot learner on standard benchmark datasets, its performance deteriorates considerably on new and unseen dialectal speech. We also notice that an MSA finetuned model does 
\textit{not} do well neither on accented nor dialectal conditions compared to a zero-shot counterpart. Further, we find that adding a language model during decoding to a small pretrained model such as XLS-R helps it outperform a whisper model of roughly the same size that is trained on 7x more Arabic data. As a future direction for our work, we intend to explore building ASR models that are robust to new unseen dialects and conditions. 

\bibliographystyle{IEEEtran}
\bibliography{mybib}

\begin{thebibliography}{10}
\providecommand{\url}[1]{#1}
\csname url@samestyle\endcsname
\providecommand{\newblock}{\relax}
\providecommand{\bibinfo}[2]{#2}
\providecommand{\BIBentrySTDinterwordspacing}{\spaceskip=0pt\relax}
\providecommand{\BIBentryALTinterwordstretchfactor}{4}
\providecommand{\BIBentryALTinterwordspacing}{\spaceskip=\fontdimen2\font plus
\BIBentryALTinterwordstretchfactor\fontdimen3\font minus
  \fontdimen4\font\relax}
\providecommand{\BIBforeignlanguage}[2]{{%
\expandafter\ifx\csname l@#1\endcsname\relax
\typeout{** WARNING: IEEEtran.bst: No hyphenation pattern has been}%
\typeout{** loaded for the language `#1'. Using the pattern for}%
\typeout{** the default language instead.}%
\else
\language=\csname l@#1\endcsname
\fi
#2}}
\providecommand{\BIBdecl}{\relax}
\BIBdecl

\bibitem{liu2022audio}
S.~Liu, A.~Mallol-Ragolta, E.~Parada-Cabaleiro, K.~Qian, X.~Jing, A.~Kathan,
  B.~Hu, and B.~W. Schuller, ``Audio self-supervised learning: A survey,''
  \emph{Patterns}, vol.~3, no.~12, p. 100616, 2022.

\bibitem{schiappa2022self}
M.~C. Schiappa, Y.~S. Rawat, and M.~Shah, ``Self-supervised learning for
  videos: A survey,'' \emph{ACM Computing Surveys}, 2022.

\bibitem{qiu2020pre}
X.~Qiu, T.~Sun, Y.~Xu, Y.~Shao, N.~Dai, and X.~Huang, ``Pre-trained models for
  natural language processing: A survey,'' \emph{Science China Technological
  Sciences}, vol.~63, no.~10, pp. 1872--1897, 2020.

\bibitem{kim22kinterspeech}
E.~Kim, J.-J. Jeon, H.~Seo, and H.~Kim, ``{Automatic Pronunciation Assessment
  using Self-Supervised Speech Representation Learning},'' in \emph{Proc.
  Interspeech 2022}, 2022, pp. 1411--1415.

\bibitem{radford2022robust}
A.~Radford, J.~W. Kim, T.~Xu, G.~Brockman, C.~McLeavey, and I.~Sutskever,
  ``Robust speech recognition via large-scale weak supervision,'' \emph{arXiv
  preprint arXiv:2212.04356}, 2022.

\bibitem{babu2021xls}
A.~Babu, C.~Wang, A.~Tjandra, K.~Lakhotia, Q.~Xu, N.~Goyal, K.~Singh, P.~von
  Platen, Y.~Saraf, J.~Pino \emph{et~al.}, ``Xls-r: Self-supervised
  cross-lingual speech representation learning at scale,'' \emph{arXiv preprint
  arXiv:2111.09296}, 2021.

\bibitem{wang2019overview}
D.~Wang, X.~Wang, and S.~Lv, ``An overview of end-to-end automatic speech
  recognition,'' \emph{Symmetry}, vol.~11, no.~8, p. 1018, 2019.

\bibitem{panayotov2015librispeech}
V.~Panayotov, G.~Chen, D.~Povey, and S.~Khudanpur, ``Librispeech: an asr corpus
  based on public domain audio books,'' in \emph{2015 IEEE international
  conference on acoustics, speech and signal processing (ICASSP)}.\hskip 1em
  plus 0.5em minus 0.4em\relax IEEE, 2015, pp. 5206--5210.

\bibitem{rousseau2012ted}
A.~Rousseau, P.~Del{\'e}glise, and Y.~Esteve, ``Ted-lium: an automatic speech
  recognition dedicated corpus.'' in \emph{LREC}, 2012, pp. 125--129.

\bibitem{ardila2019common}
R.~Ardila, M.~Branson, K.~Davis, M.~Henretty, M.~Kohler, J.~Meyer, R.~Morais,
  L.~Saunders, F.~M. Tyers, and G.~Weber, ``Common voice: A
  massively-multilingual speech corpus,'' \emph{arXiv preprint
  arXiv:1912.06670}, 2019.

\bibitem{olivier2022there}
R.~Olivier and B.~Raj, ``There is more than one kind of robustness: Fooling
  whisper with adversarial examples,'' \emph{arXiv preprint arXiv:2210.17316},
  2022.

\bibitem{ahmed2019end}
A.~Ahmed, Y.~Hifny, K.~Shaalan, and S.~Toral, ``End-to-end lexicon free arabic
  speech recognition using recurrent neural networks,'' in \emph{Computational
  Linguistics, Speech And Image Processing For Arabic Language}.\hskip 1em plus
  0.5em minus 0.4em\relax World Scientific, 2019, pp. 231--248.

\bibitem{belinkov2019analyzing}
Y.~Belinkov, A.~Ali, and J.~Glass, ``Analyzing phonetic and graphemic
  representations in end-to-end automatic speech recognition,'' \emph{arXiv
  preprint arXiv:1907.04224}, 2019.

\bibitem{hussein2022arabic}
A.~Hussein, S.~Watanabe, and A.~Ali, ``Arabic speech recognition by end-to-end,
  modular systems and human,'' \emph{Computer Speech \& Language}, vol.~71, p.
  101272, 2022.

\bibitem{baevski2020Wav2vec}
A.~Baevski, Y.~Zhou, A.~Mohamed, and M.~Auli, ``wav2vec 2.0: A framework for
  self-supervised learning of speech representations,'' \emph{Advances in
  neural information processing systems}, vol.~33, pp. 12\,449--12\,460, 2020.

\bibitem{hsu2021hubert}
W.-N. Hsu, B.~Bolte, Y.-H.~H. Tsai, K.~Lakhotia, R.~Salakhutdinov, and
  A.~Mohamed, ``Hubert: Self-supervised speech representation learning by
  masked prediction of hidden units,'' \emph{IEEE/ACM Transactions on Audio,
  Speech, and Language Processing}, vol.~29, pp. 3451--3460, 2021.

\bibitem{chung2021w2v}
Y.-A. Chung, Y.~Zhang, W.~Han, C.-C. Chiu, J.~Qin, R.~Pang, and Y.~Wu,
  ``W2v-bert: Combining contrastive learning and masked language modeling for
  self-supervised speech pre-training,'' in \emph{2021 IEEE Automatic Speech
  Recognition and Understanding Workshop (ASRU)}.\hskip 1em plus 0.5em minus
  0.4em\relax IEEE, 2021, pp. 244--250.

\bibitem{babu22interspeech}
A.~Babu, C.~Wang, A.~Tjandra, K.~Lakhotia, Q.~Xu, N.~Goyal, K.~Singh, P.~{von
  Platen}, Y.~Saraf, J.~Pino, A.~Baevski, A.~Conneau, and M.~Auli, ``{XLS-R:
  Self-supervised Cross-lingual Speech Representation Learning at Scale},'' in
  \emph{Proc. Interspeech 2022}, 2022, pp. 2278--2282.

\bibitem{valk2021voxlingua107}
J.~Valk and T.~Alum{\"a}e, ``Voxlingua107: a dataset for spoken language
  recognition,'' in \emph{2021 IEEE Spoken Language Technology Workshop
  (SLT)}.\hskip 1em plus 0.5em minus 0.4em\relax IEEE, 2021, pp. 652--658.

\bibitem{bakheet2021improving}
M.~Bakheet, ``Improving speech recognition for arabic language using low
  amounts of labeled data,'' 2021.

\bibitem{conneau2022FLEURS}
A.~Conneau, M.~Ma, S.~Khanuja, Y.~Zhang, V.~Axelrod, S.~Dalmia, J.~Riesa,
  C.~Rivera, and A.~Bapna, ``Fleurs: Few-shot learning evaluation of universal
  representations of speech,'' \emph{arXiv preprint arXiv:2205.12446}, 2022.

\bibitem{ali2016mgb}
A.~Ali, P.~Bell, J.~Glass, Y.~Messaoui, H.~Mubarak, S.~Renals, and Y.~Zhang,
  ``The mgb-2 challenge: Arabic multi-dialect broadcast media recognition,'' in
  \emph{2016 IEEE Spoken Language Technology Workshop (SLT)}.\hskip 1em plus
  0.5em minus 0.4em\relax IEEE, 2016, pp. 279--284.

\bibitem{ali2017speech}
A.~Ali, S.~Vogel, and S.~Renals, ``Speech recognition challenge in the wild:
  Arabic mgb-3,'' in \emph{2017 IEEE Automatic Speech Recognition and
  Understanding Workshop (ASRU)}.\hskip 1em plus 0.5em minus 0.4em\relax IEEE,
  2017, pp. 316--322.

\bibitem{ali2019mgb}
A.~Ali, S.~Shon, Y.~Samih, H.~Mubarak, A.~Abdelali, J.~Glass, S.~Renals, and
  K.~Choukri, ``The mgb-5 challenge: Recognition and dialect identification of
  dialectal arabic speech,'' in \emph{2019 IEEE Automatic Speech Recognition
  and Understanding Workshop (ASRU)}.\hskip 1em plus 0.5em minus 0.4em\relax
  IEEE, 2019, pp. 1026--1033.

\bibitem{chowdhury21interspeech}
S.~A. Chowdhury, A.~Hussein, A.~Abdelali, and A.~Ali, ``{Towards One Model to
  Rule All: Multilingual Strategy for Dialectal Code-Switching Arabic ASR},''
  in \emph{Proc. Interspeech 2021}, 2021, pp. 2466--2470.

\bibitem{loshchilov2017decoupled}
I.~Loshchilov and F.~Hutter, ``Decoupled weight decay regularization,''
  \emph{arXiv preprint arXiv:1711.05101}, 2017.

\end{thebibliography}
\end{document}